\documentclass{article} 
\usepackage{iclr2021_conference,times}

\usepackage{amsmath,amsfonts,bm}

\def\eqref#1{equation~\ref{#1}}

\def\1{\bm{1}}

\DeclareMathAlphabet{\mathsfit}{\encodingdefault}{\sfdefault}{m}{sl}
\SetMathAlphabet{\mathsfit}{bold}{\encodingdefault}{\sfdefault}{bx}{n}

\usepackage{hyperref}
\usepackage{url}
\usepackage{breqn}
\usepackage{amsthm}
\usepackage{amsmath}

\title{UNA: A Unified Supervised Framework for Efficient LLM Alignment Across Feedback Types}

\author{$\textbf{Zhichao Wang}^{\dagger*}, \textbf{Bin Bi}^{\dagger}$ \\
  Salesforce \\
  \texttt{\{zhichaowang, bin.bi\}@salesforce.com}
  \And
  Can Huang \\
  School of Mathematical Sciences \\
  Xiamen University \\
  \texttt{canhuang@xmu.edu.cn} \\
  \And
  Shiva Kumar Pentyala, Zixu (James) Zhu, Sitaram Asur, Na (Claire) Cheng \\
  Salesforce \\
  \texttt{\{shivakumar.pentyala,james.zhu,sasur,claire.cheng\}@salesforce.com}
  \And
  Cheng Wan \\
  RadixArk \\
  \texttt{cheng.wan@radixark.ai}
  \And
  Dong Nie \\
  ChatAlpha AI \\
  \texttt{dongnie@cs.unc.edu}
  \And
  Lingzi Hong \\
  University of North Texas \\
  \texttt{lingzi.hong@unt.edu}
}
\iclrfinalcopy 
\begin{document}

\maketitle
\def\thefootnote{}\footnotetext{*: corresponding author; $\dagger$: equal contribution}
\def\thefootnote{}\footnotetext{Code can be found at: \href{https://github.com/ZhichaoWang970201/UNA-UFT/}{https://github.com/ZhichaoWang970201/UNA-UFT/}}

\begin{abstract}
RL alignment methods, including RLHF and DPO, are primarily based on pairwise preference data. Although scalar or score-based feedback has been collected in some settings, it is rarely used directly, and preference magnitude information is typically ignored. Furthermore, current alignment frameworks offer limited capability for unifying heterogeneous supervision signals, making it difficult to jointly leverage diverse data types within a single training paradigm. This limitation constrains the richness and scalability of the alignment process. To address this gap, we propose a \textbf{UN}ified \textbf{A}lignment (UNA) framework capable of training across different types of feedback, including binary, pairwise, and score-based, through a generalized implicit reward function. The reward function is theoretically proved to be the optimal policy by the log sum inequality. Extensive experiments on classical benchmarks consistently demonstrate the advantage of the proposed unified framework with typical LLM base models.
\end{abstract}

$\textbf{\textit{Keywords: }} \text{LLM Alignment} \cdot \text{Unified Alignment} \cdot \text{RLHF} \cdot \text{PPO} \cdot \text{DPO} \cdot \text{KTO}$

\section{Introduction}
LLMs are deployed in diverse, real-world settings where feedback is rarely uniform. Therefore, it is important to achieve alignment training across different types of supervision signals.
Heterogeneous data, such as the pairwise preference data, score feedback, Likert-scale human ratings, and domain-specific evaluation metrics, capture different facets of human expectations. 
A framework that can unify these heterogeneous data sources enables models to leverage substantially richer information and reduces reliance on any single annotation paradigm. 

Score-based feedback encodes the degree or intensity of preference, offering more fine-grained guidance than binary comparisons. 
Several datasets already provide such score signals. For example, 
OpenAI’s WebGPT~\cite{nakano2021webgpt} includes Likert-scale human ratings of quality; 
AnthropicHH~\footnote{https://huggingface.co/datasets/Anthropic/hh-rlhf} dataset has multi-level preference labels;
UltraFeedback~\cite{cui2023ultrafeedback} provides 0–10 quality ratings across multiple dimensions;
and 
HelpSteer~\cite{wang2024helpsteer} and HelpSteer 2~\cite{wang2024helpsteer2} include numeric helpfulness and safety scores. 
These datasets include informative supervision signals, yet few alignment methods can utilize score-based feedback. 

Especially, there is no unified framework that integrates diverse forms of feedback. Existing methods often require separate training pipelines for each feedback type, e.g., Reinforcement Learning from Human Feedback (RLHF) \cite{ouyang2022training} with PPO for rewards, Direct Preference Optimization (DPO) \cite{rafailov2023direct} for pairwise data. Little work has been done to unify diverse forms of feedback into a single training objective. This allows training to proceed under one consistent optimization framework and has the potential to utilize a broader supervision space for more robust and effective alignment.

\begin{figure*}[ht]
    \centering
    \includegraphics[width=\textwidth]{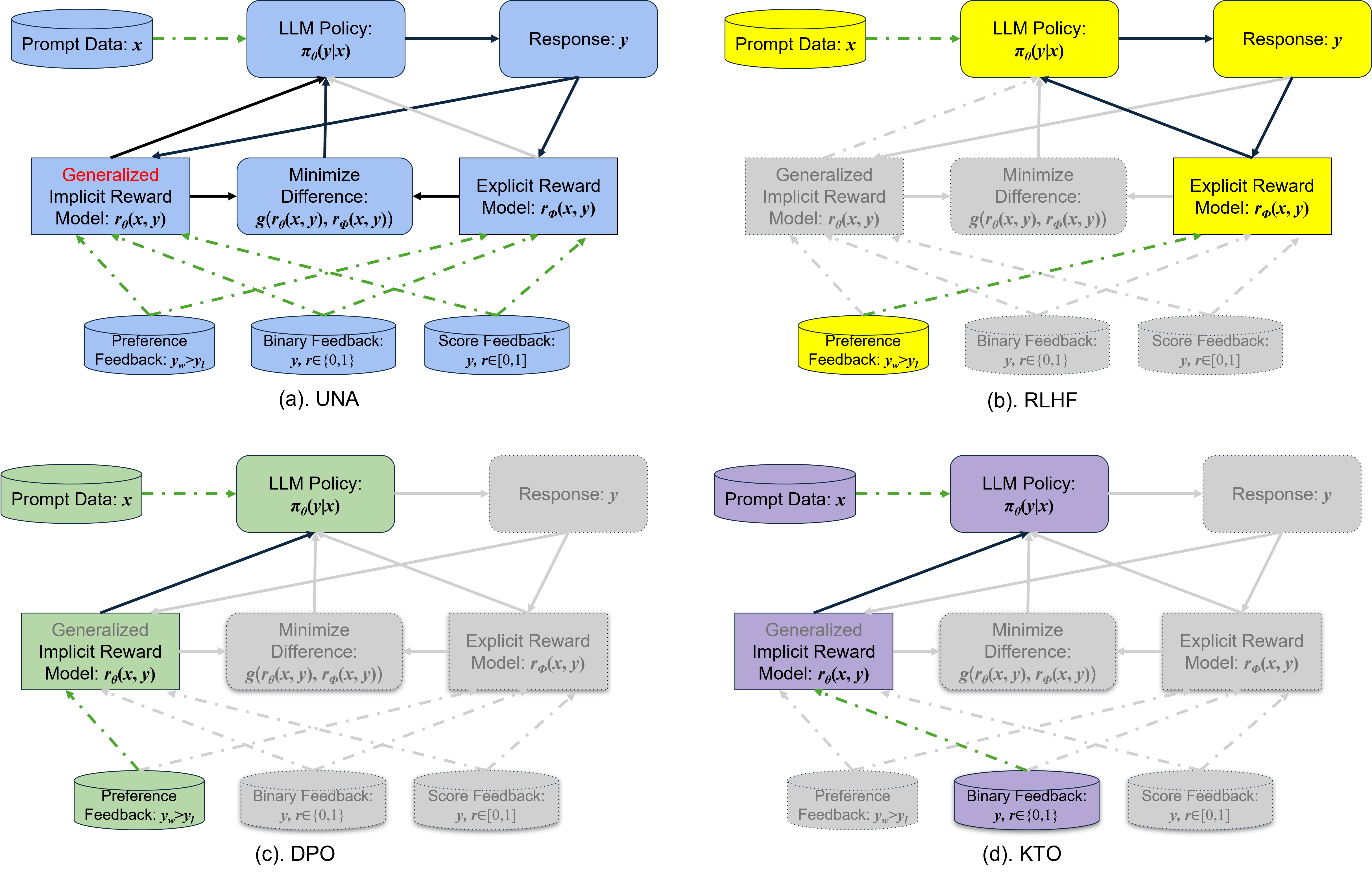}
    \caption{A figure comparison among (a). UNA, (b) RLHF, (c) DPO and (d) KTO. Each subfigure is composed of four types of data: ``prompt data", ``preference feedback", ``binary feedback" and ``score feedback", ``LLM policy", ``response", two reward models: ``generalized implicit reward model" and ``explicit reward model" and a module to minimize the difference between implicit and explicit rewards. The connection between data to other modules are utilizing green dash arrow, while others are connected by black solid arrow. All unused modules are grayed out. In part (b), RLHF firstly utilizes preference feedback to train the explicit reward model, and then use the evaluation provided by the explicit reward model to continuous optimize the policy in a online mode. In comparison, in part (c) and (d), DPO and KTO utilize preference feedback and binary feedback respectively to generate implicit reward to align LLM policy. However, in part (a), UNA can utilize \textbf{different types of data} to get the generalized implicit and explicit rewards and minimize their differences to align LLM policy in \textbf{both online and offline modes}.
}
    \label{fig:RLHF/PPO-DPO-KTO-UNA}
\end{figure*}

RLHF, as shown in Figure \ref{fig:RLHF/PPO-DPO-KTO-UNA}(b), is a two-stage process. First, a Reward Model (RM) is trained on pairwise preference data. Next, the LLM policy is fine-tuned through RL, typically using proximal policy optimization (PPO) \cite{schulman2017proximal}, where the RM evaluates the generated responses. However, RLHF faces several limitations: overfitting in RM training \cite{zhu2024iterative,huyen2023rlhf}, unstable RL fine-tuning \cite{ma2024coevolving,byun2024ares}, and high memory requirements for maintaining multiple models (policy, reference policy, RM, and value model) \cite{wang2024reinforcement}.

DPO simplifies this by mathematically establishing a mapping between the RM and the optimal policy, combining the RM and RL training into a single, stable binary classification problem (Figure \ref{fig:RLHF/PPO-DPO-KTO-UNA}(c)). This eliminates the need for an explicit RM and reduces memory costs. The policy is optimized using the difference in implicit rewards calculated for desired and undesired responses. Kahneman-Tversky Optimization (KTO) \cite{ethayarajh2024kto} extends DPO to use binary feedback for desired and undesired responses (Figure \ref{fig:RLHF/PPO-DPO-KTO-UNA}(d)). Despite their efficiency, both DPO and KTO require labeled training data for direct supervised optimization, whereas RLHF/PPO relies on a learned RM to guide policy optimization.


We propose UNA, a method capable of training with different data types, addressing the limitations of RLHF/PPO, DPO, and KTO. 
We design an implicit reward model in UNA and prove that the optimal policy $\pi^*_\theta(y|x)$ for the RLHF objective is achieved when the implicit reward model is satisfied.
Based on this generalized implicit reward, UNA unifies RLHF/PPO, DPO, and KTO into a supervised learning framework that minimizes the difference between an implicit reward and an explicit reward (Figure \ref{fig:RLHF/PPO-DPO-KTO-UNA}(a)). The explicit reward can be derived from human labelers, reward functions, or LLMs. 
UNA reformulates alignment by replacing PPO-based reinforcement learning in conventional RLHF pipelines with a supervised learning objective that minimizes the differences between an implicit reward derived from the policy and an explicit reward signal, enabling more stable and computationally efficient training.

This paper has \textbf{the following contributions}:

(1) We propose UNA, a unified alignment framework that reformulates RLHF/PPO, DPO, and KTO within a supervised learning paradigm, which can flexibly accommodate diverse feedback types, including pairwise, binary, and scalar rewards, under both online and offline training settings.

(2) We mathematically prove that the optimal policy derived from the RLHF objective function is induced by the reward function \(r_\theta(x, y) = \beta \log \left( \frac{\pi_{\theta}(y | x)}{\pi_{\text{ref}}(y | x)}\right)\).

(3) We conduct comprehensive experiments across multiple benchmark datasets and show that UNA surpasses DPO and KTO across different tasks. It achieves this while simplifying the RL fine-tuning process in RLHF/PPO, improving stability, and significantly reducing memory overhead.

\section{Related Work}
The LLM field has been transformed by large-scale pretraining with billions of parameters and trillions of tokens \cite{openai2024gpt4, anthropic2024claude, team2023gemini}. After pretraining, SFT is applied to enhance performance on downstream tasks. However, pretraining and SFT alone cannot resolve bias and ethical issues inherent in the training data \cite{openai2024gpt4}.

To address these challenges, RLHF with PPO has been widely adopted to align LLMs, including GPT and Claude \cite{ouyang2022training, bai2022training}. Despite its effectiveness, RLHF/PPO suffers from high memory usage, instability, and multi-stage training complexity, including separate RM training and RL fine-tuning \cite{rafailov2023direct}. To reduce the cost of human labeling, AI feedback can replace human feedback in a method called RLAIF \cite{bai2022constitutional, lee2023rlaif}. RLOO, in contrast, considers PPO overkill for pretrained LLMs and provides a simpler alignment alternative \cite{ahmadian2024basicsrevisitingreinforcestyle}. 

RLHF and RLAIF remain complex, unstable, and memory-intensive; GRPO addresses these issues by removing the PPO value model and using the average reward as the advantage baseline \cite{shao2024deepseekmathpushinglimitsmathematical}. In DAPO \cite{yu2025dapoopensourcellmreinforcement}, the authors utilized (i) a higher ceiling clip, (ii) dynamic sampling, (iii) token-level policy gradient loss, and (iv) overlong reward shaping to improve the stability of GRPO. GSPO replaces (i) the token-level importance ratio with a sequence-level importance ratio and (ii) routing replay to stabilize the expert choice in GRPO \cite{zheng2025groupsequencepolicyoptimization}.

\noindent \textbf{Simplifying RLHF: DPO and Variants}
DPO simplifies RLHF by mapping the optimal policy to the reward model in a single step, transforming unstable RL training into a binary classification problem \cite{rafailov2023direct}. DPOP \cite{pal2024smaug} mathematically demonstrates that the reward of desired responses may decrease during DPO and introduces a maximum term to prevent this. IPO identifies that under nearly deterministic conditions, the KL divergence constraint imposed by $\beta$ may become ineffective, potentially leading to overfitting, and proposes a new loss term to mitigate this \cite{azar2023general}. Sequential DPO (sDPO) divides the dataset into splits and aligns the model sequentially, achieving better performance than using the entire dataset at once \cite{kim2024sdpo}. Iterative DPO leverages the LLM as both response generator and evaluator to iteratively improve itself \cite{yuan2024selfrewardinglanguagemodels, xu2024thingscringeothersiterative}. TDPO provides token-level rewards to refine generation \cite{rafailov2024rqlanguagemodel, zeng2024tokenleveldirectpreferenceoptimization}.

\noindent \textbf{Merging SFT with Alignment}
Several works integrate SFT with alignment. For example, ORPO introduces a loss function that increases the ratio of desired over undesired responses, achieving the goals of both SFT and alignment \cite{hong2024orpomonolithicpreferenceoptimization}. PAFT conducts SFT and alignment in parallel and merges the results afterward \cite{pentyala2024paftparalleltrainingparadigm}. R-DPO \cite{park2024disentangling} and SimPO \cite{meng2024simpo} address the verbosity problem in outputs, using length-control methods to reduce response length while maintaining performance.

\noindent \textbf{Feedback Types: Pairwise, Binary, and Ranking}
Early work focused on pairwise datasets, which are costly to collect. Binary feedback, such as ``thumbs up'' or ``thumbs down,'' is easier to obtain. KTO leverages human preference between desired and undesired responses for effective binary feedback alignment \cite{ethayarajh2024kto}, while DRO optimizes binary feedback by estimating policy and value functions sequentially \cite{richemond2024offlineregularisedreinforcementlearning}. Nash learning models LLM improvement as a min-max problem, addressing intransitivity in human preferences through iterative optimization \cite{munos2024nashlearninghumanfeedback}, though at the cost of increased training time. SPPO uses a single model to simulate both sides of a competitive setup \cite{wu2024selfplaypreferenceoptimizationlanguage}.

Ranking-based approaches, such as LiPO \cite{liu2024lipolistwisepreferenceoptimization}, RRHF \cite{yuan2023rrhfrankresponsesalign}, and PRO \cite{song2024preferencerankingoptimizationhuman}, utilize the ranking of response lists and relative scores. RPO minimizes KL divergence between predicted and labeled rewards, aligning closely with the method proposed in this work \cite{nvidia2024nemotron4340btechnicalreport}.

\noindent \textbf{Open Challenges}
Despite these advances, several challenges remain: (i) No unified method currently combines RLHF and DPO; (ii) There is no unified framework that integrates diverse forms of feedback, including pairwise, binary, and listwise ranking.
These limitations motivate the development of UNA, which aims to unify and simplify LLM alignment methods such as PPO, DPO, and KTO, while improving training stability and efficiency compared to conventional RLHF pipelines. Unlike RLHF, which relies on reinforcement learning with an explicit reward optimization loop, UNA formulates alignment as a supervised learning problem. This design enables more stable training dynamics and higher computational efficiency while retaining alignment effectiveness.

\section{\textbf{UN}ified \textbf{A}lignment (UNA) Framework}
We introduce UNA, which derives a general loss function that reformulates RLHF/PPO, DPO, and KTO under a unified supervised learning framework, allowing UNA to leverage different types of feedback data. 
We further compare UNA with existing techniques and show the relationship of UNA to DPO, KTO, and RLHF/PPO.

\subsection{UNA via Implicit Reward Modeling}
\label{Section: UNA details}
Inspired by the idea of DPO, we propose a new relationship between the implicit reward model and the optimal policy for a unified alignment framework, including RLHF/PPO, DPO, and KTO on different types of data. By adhering to the same objective outlined in RLHF (Equation \ref{eq: RL objective}), we formulate a novel connection between the implicit reward function and the optimal policy, as shown in Equation \ref{eq: UNA optimal reward / policy}. The derivation can be found in Section \ref{Section: Mathematical proof of UNA} through log-sum inequality. A more general derivation of UNA, which arrives at the same result as DPO, is presented in Section \ref{appendix: generalized UNA derivation}.

\begin{equation}
\label{eq: RL objective}
\pi^*_\theta(y|x) = \arg\max_{\pi_\theta}\;
\mathbb{E}_{x\sim D}\Big[
    \mathbb{E}_{y\sim\pi_\theta(\cdot\mid x)}\big[ r_\phi(x,y) \big]
    - \beta\, D_{\mathrm{KL}}\big(\pi_\theta(\cdot\mid x)\,\|\,\pi_{\mathrm{ref}}(\cdot\mid x)\big)
\Big]
\end{equation}

\begin{equation}
\begin{aligned}
\label{eq: UNA optimal reward / policy}
r_{\theta}(x, y) = \beta \log \left(\frac{\pi_{\theta}(y|x)}{\pi_{\text{ref}}(y|x)}\right)
\end{aligned}
\end{equation}

The optimal implicit reward formulation in Equation \ref{eq: UNA optimal reward / policy} implies that we can transform the original unstable, memory-expensive RL training process into a reward function optimization problem, i.e., a stable and memory-efficient supervised learning process. The explicit rewards in the original RL training process can be derived from multiple methods in both online and offline modes, including human labeling, LLM-as-a-Judge, and a reward model. 
Eventually, the RL fine-tuning process is transformed into a general minimization problem between explicit reward $r_{\phi}(x, y)$ and implicit reward $r_{\theta}(x, y)$ as shown in Equation \ref{eq: general loss for UNA-reward} where $g(x_{1}, x_{2})$ refers to a general function that measure the difference between $x_{1}$ and $x_{2}$ like MSE. 

The implicit and explicit rewards may be defined on different numerical scales. For example, the explicit reward can be provided on a scale of $[0,1,2,3,4,5]$, while the implicit reward is $[-1,1]$. To ensure meaningful comparison and stable optimization, rewards are normalized before training so that implicit and explicit rewards are aligned on a common scale before the difference is calculated. 

\begin{equation}
\label{eq: general loss for UNA-reward}
L_{\text{UNA-reward}}(\pi_{\theta}) = \mathbb{E}_{(x,y) \sim D} [g(r_{\phi}(x, y), r_{\theta}(x, y))]
\end{equation}

\begin{figure*}[ht]
    \centering
    \includegraphics[width=\textwidth]{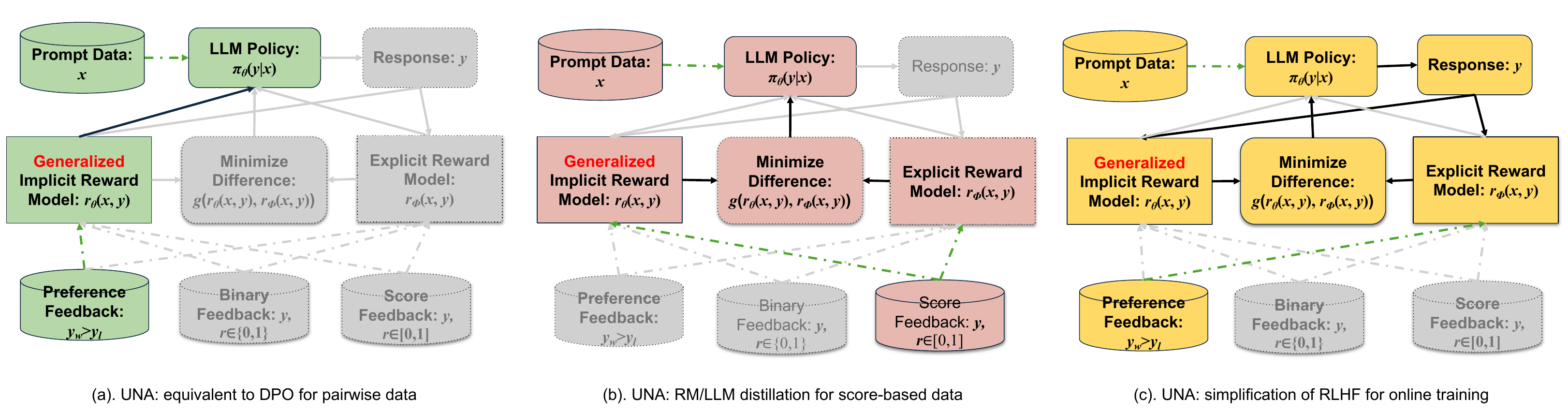}
    \caption{The two applications of UNA: Offline UNA and Online UNA. Offline UNA includes (a). equivalent to DPO for pairwise data, (b). RM/LLM-as-a-Judge distillation for score-based data. Online UNA includes (c). simplification of RLHF for online training. The same modules are utilized as in Figure \ref{fig:RLHF/PPO-DPO-KTO-UNA}, and unused modules are grayed out. For part (a), the same steps as DPO will be utilized. For part (b), (c), from the different types of data, including pairwise, binary, and score-based feedback, implicit and explicit rewards are firstly gathered. Then, the difference between implicit and explicit rewards is minimized to align the LLM policy.}
    \label{fig:UNA four applications}
\end{figure*}

Leveraging this general implicit reward function, UNA can be applied in both online and offline modes. Figure \ref{fig:UNA four applications} illustrates UNA's applications to various data types and its simplification of RLHF.






\subsubsection{Offline UNA}
In Offline UNA, prompts, responses, and their corresponding explicit rewards $(x, y, r)$ are gathered before training. These explicit rewards can encompass pairwise feedback, binary feedback, and score-based feedback, all of which the UNA framework is designed to handle seamlessly.
Offline UNA encompasses: 
(i) equivalence to DPO for pairwise preference dataset;
(ii) compatibility with binary feedback, and (iii) accommodation of score-based feedback.

\paragraph{Equivalence to DPO for Pairwise Datasets}
For pairwise datasets, the implicit rewards of desired and undesired responses can be derived as shown in Figure \ref{fig:UNA four applications}(a). Then, the LLM policy is aligned by maximizing the difference of implicit rewards between desired and undesired responses. It is equivalent to DPO as the loss function is the same as long as $g(x)=\log[\sigma(x)]$ is applied to the difference of implicit rewards of desired and undesired responses in Equation \ref{eq: general loss for UNA-reward}.

\paragraph{Compatibility with Binary Feedback}
For binary feedback, the positive and negative feedback can be transformed into explicit scores. Positive or 'thumbs up' data can be assigned an explicit reward score of 1, i.e., $r_{\phi}(x, y_w)=1$. In contrast, negative or 'thumbs down' data can be assigned an explicit reward score of 0, i.e., $r_{\phi}(x, y_l)=0$. Afterward, the implicit reward is first estimated, and then its difference from the explicit reward model is minimized on a pointwise basis, which contrasts with preference-based feedback.
Because the explicit feedback is binary, a normalization function should be utilized on the implicit reward function beforehand. Considering the implicit and explicit rewards, multiple loss functions can be formulated, like MSE and BCE. 

\paragraph{Accommodation of Score-based Feedback}
Researchers have utilized LLM-as-a-Judge and RM to evaluate responses by outputting score-based rewards according to predefined standards. If the score-based evaluations are accurate enough, they can be an extra information to utilize for alignment, compared with binary or preference feedback. When the tuple type of data (prompt, response, explicit reward) is provided, the prompt and response are utilized to calculate implicit reward as shown in Equation \ref{eq: UNA optimal reward / policy}, and the model is aligned by minimizing the difference between implicit and explicit rewards as shown in Figure \ref{fig:UNA four applications}(b). Because the explicit rewards from RM and LLM are not binary, usually a score in the interval [0, 1]. As a result, only MSE can be used as the loss function, excluding BCE. In particular, when LLM-as-a-Judge is utilized for evaluation, it can be regarded as an offline version of RLAIF.

\subsubsection{Online UNA}
Online UNA generates responses $y$ on-the-fly from the current policy given sampled prompts $x$, and computes rewards $r$ based on the resulting $(x, y)$ pairs via a learned RM. This framework aligns closely with RLHF but offers a more streamlined and stable approach to the RLHF process. Online UNA features improvement over RLHF in the RL fine-tuning stage by replacing PPO with a supervised learning process. 

\noindent\textbf{Simplification of RLHF}
When utilizing a reward model for online evaluation, UNA greatly simplifies the RL fine-tuning stage of RLHF/PPO as shown in Figure \ref{fig:UNA four applications}(c). Assuming the reward model has already been trained, the focus shifts exclusively to the RL fine-tuning stage. Prompts are first sent to the current policy for online response generation and implicit reward estimation. Then, the prompt and response are sent to the reward model for explicit reward estimation. The last step minimize the differences between implicit and explicit rewards to align the LLM policy. Eventually, the original RL objective in Equation \ref{eq: RL objective} can be transformed to difference minimization, like the MSE of implicit reward and explicit reward.

UNA has several benefits over PPO in the RL fine-tuning stage. First, it transforms the original unstable RL problem into a stable supervised learning problem by minimizing the difference between implicit and explicit rewards. Second, UNA removes the necessity of a value model in PPO, and partially reduces memory cost. Finally, the computation cost of MSE is much lower compared with the multiple terms in PPO to maintain performance. As a result, UNA will speed up the training process. 

\subsection{Theoretical Relationship of UNA to DPO and RLHF}
\noindent\textbf{UNA and DPO} The implicit rewards of UNA is presented in Equation~\ref{eq: UNA optimal reward / policy}, DPO in Equation~\ref{eq: DPO equation}. The key difference between them lies in the presence of the $\beta \log Z(x)$ term. Specifically, the implicit reward used in UNA can be viewed as a special case of DPO, where the partition function $Z(x) = \sum_{y} \pi_{\text{ref}}(y|x) \exp\left(\frac{1}{\beta} r_\theta(x, y)\right)$ is equal to 1. This condition is exactly satisfied when the reward function takes the form $r_\theta(x, y) = \beta \log \left(\frac{\pi_{\theta}(y|x)}{\pi_{\text{ref}}(y|x)}\right)$. Therefore, the optimal implicit reward function in UNA can be interpreted as a special case—and a strict subset—of the optimal implicit reward function used in DPO.

\begin{equation}
\label{eq: DPO equation}
r_\theta(x, y) = \beta \log \left(\frac{\pi_{\theta}(y|x)}{\pi_{\text{ref}}(y|x)}\right) + \beta \log Z(x)
\end{equation}

Since $Z(x)$ is generally intractable, DPO sidesteps this issue by employing pairwise preference comparisons, which allow the partition term to cancel out. However, this approach precludes the use of pointwise preference data, which often carries richer information. In contrast, UNA avoids the need for $Z(x)$ altogether, enabling effective utilization of pointwise preference signals. 
RLHF typically utilizes pointwise preference signals from a pretrained reward model. From this perspective, UNA unifies RLHF and DPO by bridging their underlying data types—pointwise and pairwise—within a common theoretical framework.

\noindent\textbf{UNA and RLHF} Both UNA and RLHF leverage pointwise rewards for aligning language models. RLHF follows a reinforcement learning paradigm, aiming to directly maximize the total explicit reward through policy optimization. 
In contrast, UNA treats the pointwise reward as an explicit target and aligns the model via supervised learning by minimizing the discrepancy between the implicit reward (induced by the policy) and the explicit reward function.
Notably, DPO also employs supervised learning but relies on pairwise preference data for alignment. From this perspective, UNA unifies RLHF and DPO by bridging their training paradigms: It adopts a supervised learning similar to DPO, while retaining the ability to leverage pointwise reward signals as in RLHF.

In summary, DPO aligns LLMs using pairwise preferences via supervised learning, while RLHF aligns LLMs using pointwise preferences through RL. UNA serves as a unifying framework that employs supervised learning to integrate both pairwise and pointwise preference data.

\section{Mathematical Proof of UNA}
\label{Section: Mathematical proof of UNA}
Here we rigorously prove that $r(x, y) = \beta \log \left(\frac{\pi_{\theta}(y|x)}{\pi_{\text{ref}}(y|x)}\right)$ will maximize the objective in RL in Equation \ref{eq: RL objective}. The proof deriving the mapping between the optimal policy and the reward model in DPO is provided in the appendix \ref{appendix: DPO: Relationship between optimal policy and reward function}.

\textbf{Proposition 1. Log Sum Inequality} Let $a_1, \ldots, a_n$ and $b_1, \ldots, b_n$ be non-negative numbers. Denote the sum of all $a_i$ by $a$, i.e., $\sum_{i=1}^n a_i=a$ and the sum of all $b_i$ by $b$, i.e., $\sum_{i=1}^n b_i=b$. The log sum inequality states Equation \ref{eq: log-sum inequality} with equality if and only if $\frac{a_i}{b_i}$ are equal for all $i$, in other words $a_i = \lambda \times b_i$ for all $i$. The proof could be found in appendix \ref{appendix: log-sum inequality}.

\begin{equation}
\label{eq: log-sum inequality}
\sum_{i=1}^n a_i \log \frac{a_i}{b_i} \geq a \log \frac{a}{b}
\end{equation}

Starting from the same objective in Equation \ref{eq: RL objective}, it can be simplified as shown in Equation \ref{eq: UNA optimal policy derivation (simplificatiion)}.

\begin{align}
\label{eq: UNA optimal policy derivation (simplificatiion)}
\begin{split}
\pi^*_\theta(y|x) &= \max_{\pi_\theta}\mathbb{E}_{x \sim D} \Bigg[\mathbb{E}_{y \sim \pi_{\theta}(y | x)} r_\theta(x, y) - \beta D_{\text{KL}} \left( \pi_{\theta}(y|x) \| \pi_{\text{ref}}(y|x) \right) \Bigg] \\
&= \max_{\pi_\theta}\mathbb{E}_{x \sim D} \Bigg[\mathbb{E}_{y \sim \pi_{\theta}(y | x)} \Bigg( r(x, y) 
- \beta \log \frac{\pi_{\theta}(y | x)}{\pi_{\text{ref}}(y | x)} \Bigg) \Bigg] \\
&= \max_{\pi_\theta}\mathbb{E}_{x \sim D}\Bigg[\mathbb{E}_{y \sim \pi_{\theta}(y | x)} \Bigg(\frac{1}{\beta}r(x, y)
- \log \frac{\pi_{\theta}(y | x)}{\pi_{\text{ref}}(y | x)}\Bigg)\Bigg] \\
&= \max_{\pi_\theta}\mathbb{E}_{x \sim D}\Bigg[\mathbb{E}_{y \sim \pi_{\theta}(y | x)} \Bigg(
-\log \frac{\pi_{\theta}(y | x)}{\pi_{\text{ref}}(y | x) e^{\frac{1}{\beta}r(x, y)}}\Bigg)\Bigg]
\end{split}
\end{align}

Based on the log-sum inequality in Equation \ref{eq: log-sum inequality}, the term can be further simplified as shown in Equation \ref{eq: UNA apply log-sum inequality} because both $\pi_{\theta}(y | x)$ and $\pi_{\text{ref}}(y | x) e^{\frac{1}{\beta}\left(r(x, y)\right)}$ are non-negative.

\begin{align}
\label{eq: UNA apply log-sum inequality}
\begin{split}
&\mathbb{E}_{x \sim D}\Bigg[\mathbb{E}_{y \sim \pi_{\theta}(y | x)} \Bigg(-\log \frac{\pi_{\theta}(y | x)}{\pi_{\text{ref}}(y | x) e^{\frac{1}{\beta}\left(r(x, y)\right)}}\Bigg)\Bigg] \\
&= - \mathbb{E}_{x \sim D} \Bigg[\sum_{y}\Bigg( \pi_{\theta}(y | x) \log \frac{\pi_{\theta}(y | x)}{\pi_{\text{ref}}(y | x) e^{\frac{1}{\beta}\left(r(x, y)\right)}} \Bigg)\Bigg] \\
&\leq - \mathbb{E}_{x \sim D} \Bigg[\left( \sum_{y} \pi_{\theta}(y | x) \right) 
\log \frac{\sum_{y} \pi_{\theta}(y | x)}{\sum_{y} \pi_{\text{ref}}(y | x) e^{\frac{1}{\beta}\left(r(x, y)\right)}} \Bigg] \\
&= - \mathbb{E}_{x \sim D}  \Bigg( 1 \log  \frac{1}{\sum_{y} \pi_{\text{ref}}(y | x) e^{\frac{1}{\beta}\left(r(x, y)\right)}} \Bigg) \\
&= \mathbb{E}_{x \sim D} \Bigg[ \log \left( \mathbb{E}_{y \sim \pi_{\text{ref}}(y | x)} e^{\frac{1}{\beta}\left(r(x, y)\right)} \right) \Bigg]
\end{split}
\end{align}

\begin{table*}[h]
\scriptsize  
\centering
\caption{Comparison of UNA with DPO, KTO considering pairwise, binary, and score-based data on different benchmarks. Best performance values are in bold.}
\begin{tabular}{lccccccc}
\hline
Method & TruthfulQA & IFEval & HellaSwag & ARC & WinoGrande & MMLU-pro & Math-Hard \\
\hline
Mistral 7B & 42.58 & 23.22 & 83.44 & 61.43 & 77.58 & 30.11 & 2.92 \\
~~~+ DPO & 44.75 & 26.30 & 84.42 & 62.88 & 79.16 & 30.41 & 2.25 \\
~~~+ KTO & 47.72 & 24.18 & 84.21 & 62.29 & 78.14 & 30.43 & 2.34 \\
~~~+ UNA-pairwise & 44.75 & 26.30 & 84.42 & 62.88 & 79.16 & 30.41 & 2.25 \\
~~~+ UNA-binary (BCE) & 48.33 & 26.49 & 84.60 & 63.14 & 79.40 & \textbf{30.73} & 2.99 \\
~~~+ UNA-score (MSE) & 55.09 & 37.25 & 84.52 & 63.23 & \textbf{80.27} & 29.72 & 2.77 \\
~~~+ UNA-score \& binary & \textbf{64.62} & \textbf{51.28} & \textbf{86.86} & \textbf{66.47} & 79.79 & 30.09 & \textbf{3.25} \\

\hline
Llama 8B & 45.16 & 13.19 & 81.78 & 58.11 & 76.87 & 32.73 & 5.66 \\
~~~+ DPO & 53.47 & 19.63 & 83.01 & 59.22 & 78.22 & 33.05 & 6.42 \\
~~~+ KTO & 55.07 & 23.24 & 83.15 & 59.13 & 77.66 & 32.86 & \textbf{6.95} \\
~~~+ UNA-pairwise & 53.47 & 19.63 & 83.01 & 59.22 & 78.22 & 33.05 & 6.42 \\
~~~+ UNA-binary (BCE) & 54.75 & 22.96 & 83.00 & 59.04 & 78.45 & 33.01 & 6.57 \\
~~~+ UNA-score (MSE) & 60.46 & \textbf{35.13} & 84.14 & 61.69 & \textbf{79.40} & \textbf{34.42} & 4.98 \\
~~~+ UNA-score \& binary & \textbf{61.24} & 27.61 & \textbf{85.40} & \textbf{62.88} & 78.93 & 34.25 & 5.66 \\
\hline

Gemma 4B & 39.73 & 27.47 & 77.48 & 58.02 & 72.69 & 27.92 & 6.72 \\
~~~+ DPO & 40.37 & 28.27 & 77.83 & 58.45 & 72.93 & 28.03 & 6.57 \\
~~~+ KTO & 39.71 & 27.66 & 77.56 & 58.28 & 72.61 & 27.98 & 6.57 \\
~~~+ UNA-pairwise & 40.37 & 28.27 & 77.83 & 58.45 & 72.93 & 28.03 & 6.57 \\
~~~+ UNA-binary (BCE) & 40.79 & 26.19 & 77.95 & 58.36 & 72.61 & 27.95 & 6.87 \\
~~~+ UNA-score (MSE) & 45.74 & \textbf{29.80} & 79.71 & \textbf{61.26} & \textbf{73.48} & \textbf{28.72} & 5.29 \\
~~~+ UNA-score \& binary & \textbf{46.69} & 27.54 & \textbf{81.21} & 59.30 & 72.14 & 28.49 & \textbf{7.40} \\

\hline
Qwen 8B & 52.19 & 42.27 & 79.71 & 67.92 & 77.03 & 47.21 & 29.46 \\
~~~+ DPO & 51.78 & 40.21 & 79.81 & 67.41 & 76.87 & 47.48 & 28.32 \\
~~~+ KTO & 52.07 & 43.19 & 79.58 & 67.66 & 76.56 & 47.18 & 28.25 \\
~~~+ UNA-pairwise & 51.78 & 40.21 & 79.81 & 67.41 & 76.87 & 47.48 & 28.32 \\
~~~+ UNA-binary (BCE) & 52.10 & 39.57 & 79.34 & 66.72 & 76.80 & 46.89 & 25.38 \\
~~~+ UNA-score (MSE) & 64.44 & 53.46 & 81.00 & \textbf{68.69} & \textbf{78.37} & \textbf{48.94} & 34.59 \\
~~~+ UNA-score \& binary & \textbf{66.92} & \textbf{64.14} & \textbf{82.92} & 68.43 & 73.72 & 44.83 & \textbf{42.60} \\
\hline
\end{tabular}
\label{Table: UNA comparison merged benchmark}
\vspace{-8pt}
\end{table*}

As a result, the maximum value of the objective function in Equation \ref{eq: RL objective} is $\beta \mathbb{E}_{x \sim D} \left\{ \log \left( \mathbb{E}_{y \sim \pi_{\text{ref}}(y | x)} e^{\frac{1}{\beta}\left(r(x, y)\right)} \right) \right\}$ in Equation \ref{eq: UNA apply log-sum inequality}, and this inequality reaches the equality condition when Equation \ref{eq: UNA optimal condition: particular} is satisfied where $\lambda$ is a constant.

\begin{equation}
\label{eq: UNA optimal condition: particular}
\frac{ \pi_{\theta}(y | x) }{ \pi_{\text{ref}}(y | x) e^{\frac{1}{\beta}r(x, y)} } = \frac{1}{\lambda}
\end{equation}

By rewriting this term, we can obtain the reward in term of the policy, i.e., $r(x, y) = \beta \log \left( \frac{\pi_{\theta}(y | x)}{\pi_{\text{ref}}(y | x)}\right) + \beta \log (\lambda)$. In special case, $\lambda = 1$, it is simplified to $r(x, y) = \beta \log \left( \frac{\pi_{\theta}(y | x)}{\pi_{\text{ref}}(y | x)} \right)$. 

A more generalized UNA derivation is presented in Appendix \ref{math proof of generalized UNA}. This derivation confirms that the same fundamental relationship between the reward model and the policy model, as defined by DPO, is maintained (Equation \ref{eq: DPO equation}). It also includes a discussion detailing the conditions $\lambda=1$.

\begin{table*}[h]
\centering
\scriptsize 
\caption{The comparison of UNA with RLHF using \texttt{HelpSteer2} prompts on different benchmarks.}
\begin{tabular}{lccccccc}
\hline
Method & TruthfulQA & IFEval & HellaSwag & ARC & WinoGrande & MMLU-Pro & Math-Hard \\ 
\hline
Qwen2-1.5B-INST & 45.93 & 22.20 & 66.72 & 43.94 & 66.06 & \textbf{25.56} & 5.40 \\
~~~+ RLHF            & 46.93 & 22.37 & 66.56 & 42.83 & 64.88 & 25.17 & \textbf{5.48} \\
~~~+ UNA             & \textbf{47.08} & \textbf{24.78} & \textbf{66.98} & \textbf{44.28} & \textbf{65.27} & 25.30 & 5.40 \\
\hline
Mistral-7B-INST & \textbf{55.94} & 38.46 & 75.99 & \textbf{55.29} & 73.72 & 24.53 & \textbf{2.02} \\
~~~+ RLHF            & 55.88 & 38.53 & 76.03 & 55.20 & 73.56 & 24.60 & 1.79 \\
~~~+ UNA             & 55.88 & \textbf{39.17} & \textbf{76.61} & 55.20 & \textbf{74.03} & \textbf{24.87} & 1.75 \\
\hline
\end{tabular}
\label{Table: UNA RLHF comparison selected}
\end{table*}
\section{Experiments and Results}
\label{Experiments}
We evaluate UNA under two experimental settings. 

\noindent \textbf{Offline Setting} \texttt{mistralai/Mistral-7B-v0.1}~\cite{jiang2023mistral7b}, \texttt{meta-llama/Llama-3.1-8B}~\cite{grattafiori2024llama3herdmodels}, \texttt{Qwen/Qwen3-8B-Base}~\cite{yang2025qwen3technicalreport}, and \texttt{google/gemma-3-4b-pt}~\cite{gemmateam2025gemma3technicalreport} are utilized as the policy model, and the \texttt{HelpSteer2} dataset \cite{nvidia2024nemotron4340btechnicalreport} is used as the alignment data, which have a prompt, chosen and rejected responses with corresponding scores that are labeled by humans from the perspectives of \emph{helpfulness}, \emph{correctness}, \emph{coherence}, \emph{complexity}, and \emph{verbosity}. The combined score is computed as: $0.65 \times \text{helpfulness} + 0.8 \times \text{correctness} + 0.45 \times \text{coherence}$, following~\cite{wang2024helpsteer2}. 
For binary feedback, the chosen responses are regarded as desired responses with reward ``+1" and rejected responses are regarded as undesired responses with reward ``0". The score-based feedback includes a rating of 0 to 4 for each metric in \texttt{HelpSteer2}. The rewards are weighted and normalized and used as explicit feedback to align the LLM.

Low rank adaptation (LoRA) \cite{hu2021loralowrankadaptationlarge} is employed during the fine-tuning process with $r=32$, where $r$ denotes the ranks used in LoRA. Beam search is used to identify the optimal combination of $\beta$ and learning rate. The selected configurations are listed.
UNA-binary uses $\beta = 0.01$, while DPO, KTO, and UNA-score utilize $\beta = 0.03$. Furthermore, UNA-score employs a learning rate of $3 \times 10^{-5}$, whereas the other methods use a learning rate of $5 \times 10^{-6}$.

\noindent \textbf{Online Setting} We use policy model \texttt{Qwen/Qwen2-1.5B-Instruct}~\cite{yang2024qwen2technicalreport} and \texttt{mistralai/Mistral-7B-Instruct}~\cite{jiang2023mistral7b}, and reward model \texttt{Ray2333/GRM-Llama3.2-3B-rewardmodel-ft} \cite{yang2024regularizing}. We use prompts from \texttt{Helpsteer2}, excluding those longer than 512 tokens. In RLHF, prompts are used for response generation, reward estimation via a reward model, and policy updates through PPO. In contrast, online UNA uses the same prompts for response generation, implicit reward estimation by the policy, explicit reward estimation by the reward model, and policy updates via discrepancy minimization (e.g., MSE) between implicit and explicit rewards.

Similarly, we identify the optimal combination of parameters with beam search. For $\beta$, RLHF utilizes 10, while UNA uses 30, with both approaches employing the same learning rate of $3\times 10^{-6}$.

After alignment, seven benchmark tasks are utilized to measure the performance, including TruthfulQA \cite{lin2022truthfulqa}, IFEval \cite{zhou2023instructionfollowingevaluationlargelanguage}, HellaSwag \cite{zellers2019hellaswag}, ARC \cite{allenai:arc}, WinoGrande \cite{sakaguchi2019winograndeadversarialwinogradschema}, MMLU-pro \cite{wang2024mmluprorobustchallengingmultitask}, and Math-Hard \cite{hendrycks2021measuringmathematicalproblemsolving}. In addition to evaluating the model's selection capabilities from predefined candidate answers, Alpaca-eval is used to assess the model's ability to generate text responses.

\subsection{Offline: Improvements over DPO \& KTO}

The results are presented in Table \ref{Table: UNA comparison merged benchmark}, along with the following insights. (1) UNA works with different forms of feedback. (2) On binary data, UNA outperforms DPO and KTO. (3) With score-based feedback, UNA outperforms UNA-binary, benefiting from the additional information provided by scalar scores. (4) Although the settings are not directly comparable due to differences in training data, UNA with both binary and score feedback outperforms the score-only variant on many tasks where binary data are from Helpsteer3 \cite{wang2025helpsteer3preferenceopenhumanannotatedpreference}.
We also conducted evaluations on AlpacaEval \cite{alpaca_eval}, and UNA-score achieves the highest performance ( seen Table \ref{Table: UNA comparison helpsteer} in the appendix~\ref{sec:alpacaeval}).

\subsection{Online: Improvement and Simplification over RLHF}

Table \ref{Table: UNA RLHF comparison selected} shows that online UNA outperforms RLHF on most tasks. 
The problem of alignment tax \cite{ouyang2022training} still exists on some tasks, as their performances decrease.  
Notably, by reformulating RLHF as a supervised learning problem and eliminating the value model, online UNA substantially reduces both memory consumption and training time. The training time for 20,000 steps with 8 80G A100 GPUs is around 8 hours for RLHF and 6.5 hours for online UNA with the same batch size.
The comparison of RLHF with UNA on AlpacaEval (Table \ref{Table: UNA RLHF comparison helpsteer} in appendix~\ref{sec:alpacaeval}). further demonstrates the benefit of UNA over RLHF.

\section{Conclusion}
We introduce UNA, a unified alignment framework that supports training with diverse forms of feedback. By introducing an implicit reward model and showing that satisfying this condition yields the RLHF-optimal policy, UNA provides a unified foundation for alignment across feedback modalities.
Through theoretical derivations and extensive empirical evaluation, we demonstrate that UNA effectively supports binary, pairwise, and score-based feedback. In particular, under score-based feedback, UNA can exploit pairwise difference information, leading to consistently better performance than DPO and KTO. Furthermore, we find that combining binary and score-based feedback yields additional performance gains over the score-only UNA variant across multiple tasks. Our experimental results also show that UNA outperforms RLHF in both effectiveness and training efficiency.
Overall, our findings suggest that UNA provides a general and practical alignment framework that overcomes key limitations of RLHF/PPO, DPO, and KTO, while enabling robust and effective learning from heterogeneous feedback signals.

\section*{Limitations}
\label{Limitations}
There are some limitations for this work. In this work, the theoretical unification of GRPO, which greatly reduces the computational overhead of PPO, within the UNA framework remains unexplored. In addition, the datasets utilized are limited to English research datasets, while more experiments on multilingual industrial-level datasets should be conducted.

\bibliography{iclr2021_conference}
\bibliographystyle{iclr2021_conference}

\appendix

\section{Default Notation}
\begin{description}
  \item[$x$:] prompt to LLM
  \item[$y_w:$] desired response
  \item[$y_l:$] undesired response
  \item[$P(y_w > y_l|x):$] the probability of desired response over undesired response
  \item[$r_{\phi}(x,y):$] the explicit reward
  \item[$r_{\theta}(x,y):$] the implicit reward
  \item[$s_{\phi}(x,y):$] the explicit score: normalized explicit reward
  \item[$s_{\theta}(x,y):$] the implicit score: normalized implicit reward
  \item[$D_{KL}:$] KL divergence
  \item[$\pi_{\theta}:$] LLM policy to be aligned
  \item[$\pi_{ref}:$] reference policy for LLM alignment
  \item[$g(\cdot):$] any function that measures the difference between implicit and explicit reward functions
\end{description}


\newpage
\section{DPO: Relationship between optimal policy and reward function}
\label{appendix: DPO: Relationship between optimal policy and reward function}

The objective of RLHF / DPO is shown in Equation \ref{eq: RL objective}. From the objective, the relationship between optimal reward and optimal policy can be derived in Equation \ref{eq: DPO equation} where $Z(x) = \sum_{y} \pi_{\text{ref}}(y|x) e^{\left(\frac{1}{\beta} r_\theta(x, y)\right)}$. The illustration for deriving DPO is shown in Equation \ref{eq: DPO optimal policy derivation}.

\begin{align}
\begin{split}
\label{eq: DPO optimal policy derivation}
\pi^*_\theta(y|x) &= \max_{\pi_\theta}\mathbb{E}_{x \sim D} \left[\mathbb{E}_{y \sim \pi_{\theta}(y | x)} r_\theta(x, y) - \beta D_{\text{KL}} \left( \pi_{\theta}(y|x) \| \pi_{\text{ref}}(y|x) \right)\right] \\
&= \max_{\pi_\theta}\mathbb{E}_{x \sim D}\left\{\mathbb{E}_{y \sim \pi_{\theta}(y | x)} \left[ r(x, y) - \beta \log \frac{\pi_{\theta}(y | x)}{\pi_{\text{ref}}(y | x)}\right]\right\} \\
&= \min_{\pi_\theta}\mathbb{E}_{x \sim D}\left\{\mathbb{E}_{y \sim \pi_{\theta}(y | x)} \left[\log \frac{\pi_{\theta}(y | x)}{\pi_{\text{ref}}(y | x)} - \frac{1}{\beta}r(x, y)\right]\right\} \\
&= \min_{\pi_\theta}\mathbb{E}_{x \sim D}\left\{\mathbb{E}_{y \sim \pi_{\theta}(y | x)} \left[\log \left(\frac{\pi_{\theta}(y | x)}{\frac{1}{Z(x)}\pi_{\text{ref}}(y | x) e^{\frac{1}{\beta}r(x, y)}}\right) - \log \left(Z(x)\right)\right]\right\} \\
&= \min_{\pi_\theta}\mathbb{E}_{x \sim D}\left\{\mathbb{E}_{y \sim \pi_{\theta}(y | x)} \left[\log \left(\frac{\pi_{\theta}(y | x)}{\frac{1}{Z(x)}\pi_{\text{ref}}(y | x) e^{\frac{1}{\beta}r(x, y)}}\right)\right] - \log \left(Z(x)\right) \right\} \\
&= \min_{\pi_\theta}\mathbb{E}_{x \sim D} \left\{D_{KL}\left(\pi_\theta(y|x) \| \frac{1}{Z(x)}\pi_{\text{ref}}(y | x) e^{\frac{1}{\beta}r(x, y)}\right) - \log \left(Z(x)\right) \right\}
\end{split}
\end{align}

The objective function is minimized when $D_{KL}\left(\pi_\theta(y|x)||\frac{1}{Z(x)}\pi_{\text{ref}}(y | x) e^{\frac{1}{\beta}r(x, y)}\right) = 0$, and this is equivalent to $\pi_\theta(y|x)=\frac{1}{Z(x)}\pi_{\text{ref}}(y | x) e^{\frac{1}{\beta}r(x, y)}$. By rewriting, the reward model can be expressed in term of the current policy as shown in Equation \ref{eq: DPO equation}.

However, the term $Z(x)$ cannot be computed as it needed to be computed by summing all candidate responses $y$. DPO avoids this problem by subtracting the rewards of desired and undesired responses $r(x, y_w) - r(x, y_l) = \beta\left[\log \left(\frac{\pi_{\theta}(y_w|x)}{\pi_{\text{ref}}(y_w|x)}\right) - \log \left(\frac{\pi_{\theta}(y_l|x)}{\pi_{\text{ref}}(y_l|x)}\right)\right]$. In addition, the authors argue ``We say that two reward functions $r(x, y)$ and $r'(x, y)$ are equivalent iff $r(x, y) -  r'(x, y) = f(x)$ for some function $f$". However, rigorous proof cannot be provided and it is only provided that $r(x, y)$ and $r'(x, y)$ induce the same optimal policy. For Lipo, $r(x, y) = \beta \log \left(\frac{\pi_{\theta}(y|x)}{\pi_{\text{ref}}(y|x)}\right)$ is directly utilized as rewards for listwise responses and KTO estimates $Z(x)$ by averaging over multiple samples.

\newpage
\section{Derivation of log-sum inequality}
\label{appendix: log-sum inequality}
\textbf{Jensen inequality.} For a real convex function \(\varphi\), numbers \(x_{1}, x_{2}, \ldots, x_{n}\) in its domain, and positive weights \(a_{i}\), Jensen's inequality can be stated as in Equation \ref{eq: jesen inequality}:
\vspace{-1pt}

\begin{equation}
\label{eq: jesen inequality}
\frac{\sum_{i=1}^n a_{i}\varphi (x_{i})}{\sum_{i=1}^n a_{i}}
\geq 
\varphi \left( \frac{\sum_{i=1}^n a_{i}x_{i}}{\sum_{i=1}^n a_{i}} \right)
\end{equation}
\vspace{-1pt}

\textbf{Proof of log-sum inequality.} 
Firstly, define $f(x) = x \log(x)$. Then, $f'(x)=1+\log(x)$ and $f''(x)=\frac{1}{x}$. For the domain $x > 0$, $f''(x) > 0$. As a result, $f(x) = x \log(x)$ is a convex function and satisfies Jensen's inequality. Then, the log-sum inequality could be derived in Equation \ref{eq: log-sum inequality derivation}.

\begin{align}
\label{eq: log-sum inequality derivation}
\begin{split}
\sum_{i=1}^n a_i \log \left(\frac{a_i}{b_i}\right) &= \sum_{i=1}^n b_i f\left(\frac{a_i}{b_i}\right) \\
&= b \sum_{i=1}^n \frac{b_i}{b} f\left(\frac{a_i}{b_i}\right) \\
&= b \frac{\sum_{i=1}^n b_i f\left(\frac{a_i}{b_i}\right)}{\sum_{i=1}^n b_i} \\
&\geq b f\left[ \frac{\sum_{i=1}^n b_i \frac{a_i}{b_i}}{\sum_{i=1}^n b_i} \right] \\
&= b f\left( \frac{a}{b} \right)
\end{split}
\end{align}

\newpage
\section{Mathematical Proof of the Generalized UNA and Its Relationship with DPO}
\label{math proof of generalized UNA}
Starting from the same objective in Equation \ref{eq: RL objective}, it can be simplified as shown in Equation \ref{eq: UNA optimal policy derivation (simplificatiion) appendix}.
\label{appendix: generalized UNA derivation}

\begin{align}
\label{eq: UNA optimal policy derivation (simplificatiion) appendix}
\begin{split}
\pi^*_\theta(y|x) &= \max_{\pi_\theta}\mathbb{E}_{x \sim D} \left[\mathbb{E}_{y \sim \pi_{\theta}(y | x)} r_\theta(x, y) - \beta D_{\text{KL}} \left( \pi_{\theta}(y|x) \| \pi_{\text{ref}}(y|x) \right)\right] \\
&= \max_{\pi_\theta}\mathbb{E}_{x \sim D}\left\{\mathbb{E}_{y \sim \pi_{\theta}(y | x)} \left[ r(x, y) - \beta \log \frac{\pi_{\theta}(y | x)}{\pi_{\text{ref}}(y | x)}\right]\right\} \\
&= \beta \max_{\pi_\theta}\mathbb{E}_{x \sim D}\left\{\mathbb{E}_{y \sim \pi_{\theta}(y | x)} \left[\frac{1}{\beta}r(x, y) - \log \frac{\pi_{\theta}(y | x)}{\pi_{\text{ref}}(y | x)}\right]\right\} \\
&= \beta \max_{\pi_\theta}\mathbb{E}_{x \sim D}\left\{\mathbb{E}_{y \sim \pi_{\theta}(y | x)} \left[-\log \left(\frac{\pi_{\theta}(y | x)}{\pi_{\text{ref}}(y | x) e^{\frac{1}{\beta}r(x, y)}}\right)\right]\right\} \\
&= \beta \max_{\pi_\theta}\mathbb{E}_{x \sim D}\left\{\mathbb{E}_{y \sim \pi_{\theta}(y | x)} \left[-\log \left(\frac{\pi_{\theta}(y | x)}{\pi_{\text{ref}}(y | x) e^{\frac{1}{\beta}\left(r(x, y)-f(x)\right)}}\right)+\frac{1}{\beta}f(x)\right]\right\} \\
&= \beta \max_{\pi_\theta}\mathbb{E}_{x \sim D}\left\{\mathbb{E}_{y \sim \pi_{\theta}(y | x)} \left[-\log \left(\frac{\pi_{\theta}(y | x)}{\pi_{\text{ref}}(y | x) e^{\frac{1}{\beta}\left(r(x, y)-f(x)\right)}}\right)\right]+\frac{1}{\beta}f(x)\right\}
\end{split}
\end{align}

Based on the log-sum inequality in Equation \ref{eq: log-sum inequality}, the term can be further simplified as shown in Equation \ref{eq: UNA apply log-sum inequality appendix} because both $\pi_{\theta}(y | x)$ and $\pi_{\text{ref}}(y | x) e^{\frac{1}{\beta}\left(r(x, y)-f(x)\right)}$ are non-negative.

\begin{align}
\label{eq: UNA apply log-sum inequality appendix}
\begin{split}
&\beta \mathbb{E}_{x \sim D}\left\{\mathbb{E}_{y \sim \pi_{\theta}(y | x)} \left[-\log \left(\frac{\pi_{\theta}(y | x)}{\pi_{\text{ref}}(y | x) e^{\frac{1}{\beta}\left(r(x, y)-f(x)\right)}}\right)\right]+\frac{1}{\beta}f(x)\right\} \\
&= \beta \mathbb{E}_{x \sim D} \left\{-\sum_{y}\left[ \pi_{\theta}(y | x) \log \left(\frac{\pi_{\theta}(y | x)}{\pi_{\text{ref}}(y | x) e^{\frac{1}{\beta}\left(r(x, y)-f(x)\right)}}\right) \right]+\frac{1}{\beta}f(x) \right\} \\
&\leq \beta \mathbb{E}_{x \sim D} \left\{ \left[- \left( \sum_{y} \pi_{\theta}(y | x) \right) \log \left( \frac{\sum_{y} \pi_{\theta}(y | x)}{\sum_{y} \pi_{\text{ref}}(y | x) e^{\frac{1}{\beta}\left(r(x, y)-f(x)\right)}} \right)\right]+\frac{1}{\beta}f(x)\right\} \\
&= \beta \mathbb{E}_{x \sim D} \left\{ \left[- 1 \log \left( \frac{1}{\sum_{y} \pi_{\text{ref}}(y | x) e^{\frac{1}{\beta}\left(r(x, y)-f(x)\right)}} \right)\right] + \frac{1}{\beta}f(x)\right\} \\
&= \beta \mathbb{E}_{x \sim D} \left\{ \log \left( \mathbb{E}_{y \sim \pi_{\text{ref}}(y | x)} e^{\frac{1}{\beta}\left(r(x, y)-f(x)\right)} \right) + \frac{1}{\beta}f(x) \right\}
\end{split}
\end{align}

As a result, the maximum value of the objective function $\max_{\pi_\theta}\mathbb{E}_{x \sim D} \left[\mathbb{E}_{y \sim \pi_{\theta}(y | x)} r_\theta(x, y) - \beta D_{\text{KL}} \left( \pi_{\theta}(y|x) \| \pi_{\text{ref}}(y|x) \right)\right]$ in Equation \ref{eq: UNA optimal policy derivation (simplificatiion) appendix} is $\beta \mathbb{E}_{x \sim D} \left\{ \log \left( \mathbb{E}_{y \sim \pi_{\text{ref}}(y | x)} e^{\frac{1}{\beta}\left(r(x, y)-f(x)\right)} \right) + \frac{1}{\beta}f(x) \right\}$ in Equation \ref{eq: UNA apply log-sum inequality appendix}, and this inequality reaches the equality condition when Equation \ref{eq: UNA optimal condition} is satisfied where $\lambda$ is a constant.

\begin{dmath}
\label{eq: UNA optimal condition}
\frac{\pi_{\theta}(y | x)}{\pi_{\text{ref}}(y | x) e^{\frac{1}{\beta}\left(r(x, y)-f(x)\right)}} = \frac{1}{\lambda}
\end{dmath}

By rewriting this term, we can obtain the reward in term of the policy as shown in Equation \ref{eq: UNA optimal relationship}. In special case, $f(x) = \beta \log(\lambda) = 0$, it is simplified to $r(x, y) = \beta \log \left( \frac{\pi_{\theta}(y | x)}{\pi_{\text{ref}}(y | x)} \right)$. The condition $f(x) = \beta \log(\lambda) = 0$ refers that implicit and explicit reward models are exactly the same.

\begin{dmath}
\begin{aligned}
\label{eq: UNA optimal relationship}
r(x, y) &= \beta \log \left( \frac{\lambda\pi_{\theta}(y | x)}{\pi_{\text{ref}}(y | x)} \right) + f(x) \\
&= \beta \log \left( \frac{\pi_{\theta}(y | x)}{\pi_{\text{ref}}(y | x)}\right) + f(x) + \beta \log (\lambda)
\end{aligned}
\end{dmath}

When plugging Equation \ref{eq: UNA optimal condition} in Equation \ref{eq: UNA apply log-sum inequality appendix}, the upper bound can be simplified into a constant $\beta\log(\lambda) + \mathbb{E}_{x \sim D}(f(x))$ as shown in Equation \ref{eq: UNA maximum objective}.

\begin{dmath}
\label{eq: UNA maximum objective}
\beta \mathbb{E}_{x \sim D} \left\{ \log \left( \mathbb{E}_{y \sim \pi_{\text{ref}}(y | x)} e^{\frac{1}{\beta}\left(r(x, y)-f(x)\right)} \right) + \frac{1}{\beta}f(x) \right\} \\
= \beta \mathbb{E}_{x \sim D} \left\{ \log \left( \mathbb{E}_{y \sim \pi_{\text{ref}}(y | x)} \frac{\lambda \pi_{\theta}(y | x)}{\pi_{\text{ref}}(y|x)} \right) + \frac{1}{\beta}f(x) \right\} \\
= \beta \mathbb{E}_{x \sim D} \left\{ \log \left( \mathbb{E}_{y \sim \pi_{\theta}(y | x)} \lambda \right) + \frac{1}{\beta}f(x) \right\} \\
= \beta \mathbb{E}_{x \sim D} \left\{ \log \left( \lambda \right) + \frac{1}{\beta}f(x) \right\} \\
= \beta\log(\lambda) + \mathbb{E}_{x \sim D}(f(x)) \\
\end{dmath}

When desired to generalize this into ``infinite dimension", another constraint needs to be added, i.e., $\sum_{y} \pi_{\text{ref}}(y | x) e^{\frac{1}{\beta}\left(r(x, y)-f(x)\right)}$ should be finite. Then, $f(x)$ is further restricted to $f(x) > \max [r(x, y)]$ with normalization on $r(x, y)$ in advance. Eventually, $\sum_{y} \pi_{\text{ref}}(y | x) e^{\frac{1}{\beta}\left(r(x, y)-f(x)\right)} < \sum_{y} \pi_{\text{ref}}(y | x) = 1$, which will be finite.

Lastly, the relationship between UNA and DPO will be established. Under the optimal condition of UNA, as defined in Eq.~\ref{eq: UNA optimal condition}, the probability $\pi_{\theta}(y | x)$ can be expressed as $\pi_{\theta}(y | x) = \frac{1}{\lambda} \pi_{\text{ref}}(y | x) e^{\frac{1}{\beta} \left(r(x, y) - f(x)\right)}$. Since $\pi_{\theta}(y | x)$ represents a valid probability distribution, it must satisfy the normalization condition $\sum_{y} \pi_{\theta}(y | x) = 1$. Consequently, this can be rewritten as shown in Eq.~\ref{eq: UNA probability rewritten}.

\begin{dmath}
\label{eq: UNA probability rewritten}
1 = \sum_{y} \pi_{\theta}(y | x) \\
= \sum_{y} \frac{1}{\lambda}\pi_{\text{ref}}(y | x) e^{\frac{1}{\beta}\left(r(x, y)-f(x)\right)} \\
= \sum_{y} \frac{\pi_{\text{ref}}(y | x) e^{\frac{1}{\beta} r(x, y)}}{\lambda e^{\frac{1}{\beta} f(x)}} \\
= \frac{\sum_{y} \pi_{\text{ref}}(y | x) e^{\frac{1}{\beta} r(x, y)}}{\lambda e^{\frac{1}{\beta} f(x)}} \\
= \frac{Z(x)}{\lambda e^{\frac{1}{\beta} f(x)}} \\
\end{dmath}

From Eq. \ref{eq: UNA probability rewritten}, we can derive $Z(x) = \lambda e^{\frac{1}{\beta} f(x)}$. When apply $\log$ on both sides, $\beta\log(Z(x))=\beta\log(\lambda e^{\frac{1}{\beta} f(x)})=f(x)+\beta\log(\lambda )$. The implicit reward function of DPO and UNA is unified: $r(x, y)=\beta \log \left( \frac{\pi_{\theta}(y | x)}{\pi_{\text{ref}}(y | x)}\right) + f(x) + \beta \log (\lambda)=\beta \log \left( \frac{\pi_{\theta}(y | x)}{\pi_{\text{ref}}(y | x)}\right) + \beta\log(Z(x))$.

From Eq.~\ref{eq: UNA probability rewritten}, we derive the expression for \( Z(x) \) as \( Z(x) = \lambda e^{\frac{1}{\beta} f(x)} \). Taking the natural logarithm on both sides yields $\beta \log(Z(x)) = \beta \log\left(\lambda e^{\frac{1}{\beta} f(x)}\right) = f(x) + \beta \log(\lambda)$.
Thus, the implicit reward function for DPO and UNA can be unified as $r(x, y) = \beta \log \left( \frac{\pi_{\theta}(y | x)}{\pi_{\text{ref}}(y | x)} \right) + f(x) + \beta \log(\lambda) = \beta \log \left( \frac{\pi_{\theta}(y | x)}{\pi_{\text{ref}}(y | x)} \right) + \beta \log(Z(x))$.

\newpage
\section{Result of UNA on Alpacaeval}
\label{sec:alpacaeval}
\begin{table*}[h]
\centering
\caption{The comparison of UNA with DPO, KTO, considering pairwise, binary, and score-based data on AlpacaEval using \texttt{HelpSteer2} as fine-tuning data}
\begin{tabular}{lclc}
\hline
Method & Alpacaeval LC WR & Method & Alpacaeval LC WR  \\ 
\hline
Mistral & 0.31 & Llama & 0.25\\
~~~+ DPO  & 3.67 & ~~~+ DPO  & 2.09 \\
~~~+ KTO & 4.46 & ~~~+ KTO & 4.17 \\
~~~+ UNA-pairwise & 3.67 & ~~~+ UNA-pairwise & 2.09 \\
~~~+ UNA-binary (BCE) & 7.41 & ~~~+ UNA-binary (BCE) & 3.96\\
~~~+ UNA-score (MSE) & \textbf{8.78} & ~~~+ UNA-score (MSE) & \textbf{7.87} \\
\hline
\end{tabular}
\label{Table: UNA comparison helpsteer}
\end{table*}

\begin{table*}[h]
\centering
\caption{The comparison of UNA with RLHF using \texttt{HelpSteer2} prompts on AlpacaEval}
\begin{tabular}{lclc}
\hline
Method & AlpacaEval LC WR & Method & AlpacaEval LC WR \\ 
\hline
Qwen2-1.5B-INST & 1.06 & Mistral-7B-INST & 10.31\\
~~~+ RLHF & 0.66 & ~~~+ RLHF & 10.15 \\
~~~+ UNA & \textbf{1.63} & ~~~+ UNA & \textbf{10.54} \\
\hline
\end{tabular}
\label{Table: UNA RLHF comparison helpsteer}
\end{table*}

\end{document}